\documentclass{article}

\usepackage{microtype}
\usepackage{graphicx}
\usepackage{subcaption}
\usepackage{booktabs} 

\usepackage{hyperref}

\newcommand{\status}{accepted}
\usepackage[\status]{icml2024}

\usepackage{amsmath}
\usepackage{amssymb}
\usepackage{mathtools}
\usepackage{amsthm}
\usepackage{enumitem}
\usepackage{tabularx}
\usepackage{multirow}
\usepackage{colortbl}
\usepackage{makecell}
\usepackage{hhline}

\newcolumntype{Y}{>{\centering\arraybackslash}X}
\definecolor{Green}{RGB}{0, 176, 85}
\definecolor{Red}{RGB}{192, 0, 0}
\definecolor{Gray}{RGB}{110, 110, 110}

\usepackage[capitalize,noabbrev]{cleveref}

\theoremstyle{plain}

\theoremstyle{definition}

\theoremstyle{remark}

\newcommand{\ma}{Requirement Generator}
\newcommand{\mb}{Model Customizer}
\newcommand{\mba}{Model Generator}
\newcommand{\mbb}{Parameter Generator}

\usepackage[textsize=tiny]{todonotes}

\icmltitlerunning{ModelGPT: Unleashing LLM's Capabilities for Tailored Model Generation}

\begin{document}
\twocolumn[
\icmltitle{ModelGPT: Unleashing LLM's Capabilities for Tailored Model Generation}



\icmlsetsymbol{equal}{*}

\begin{icmlauthorlist}
\icmlauthor{Zihao Tang}{zju}
\icmlauthor{Zheqi Lv}{zju}
\icmlauthor{Shengyu Zhang}{zju}
\icmlauthor{Fei Wu}{zju}
\icmlauthor{Kun Kuang}{zju}
\end{icmlauthorlist}

\icmlaffiliation{zju}{Department of Computer Science, Zhejiang University, Hangzhou, China}


\icmlkeywords{LLM, Hypernetwork, Model Generation}

\vskip 0.3in
]



\printAffiliations{}

\begin{abstract}
\ifthenelse{\equal{\status}{accepted}}
{
The rapid advancement of Large Language Models (LLMs) has revolutionized various sectors by automating routine tasks, marking a step toward the realization of Artificial General Intelligence (AGI). However, they still struggle to accommodate the diverse and specific needs of users and simplify the utilization of AI models for the average user. In response, we propose ModelGPT, a novel framework designed to determine and generate AI models specifically tailored to the data or task descriptions provided by the user, leveraging the capabilities of LLMs. Given user requirements, ModelGPT is able to provide tailored models at most 270x faster than the previous paradigms (\textit{e.g.} all-parameter or LoRA finetuning). Comprehensive experiments on NLP, CV, and Tabular datasets attest to the effectiveness of our framework in making AI models more accessible and user-friendly. Our code is available at \href{https://github.com/IshiKura-a/ModelGPT}{here}.
}{
The rapid advancement of Large Language Models (LLMs) has revolutionized various sectors by automating routine tasks. However, they still struggle to accommodate the diverse and specific needs of users and simplify the utilization of AI models for the average user. In response, we propose ModelGPT, a novel framework designed to determine and generate AI models specifically tailored to the data or task descriptions provided by the user, leveraging the capabilities of LLMs. Given user requirements, ModelGPT is able to provide tailored models at most 270x faster than the previous paradigms (\textit{e.g.} all-parameter or LoRA finetuning). Comprehensive experiments on NLP, CV, and Tabular datasets attest to the effectiveness of our ModelGPT in making AI models more accessible and user-friendly.
}
\end{abstract}

\section{Introduction}
In recent years, advancements in Large Language Models (LLMs) such as GPT-4 \cite{DBLP:journals/corr/abs-2303-08774}, LLaMA \cite{DBLP:journals/corr/abs-2302-13971} have significantly influenced people's daily life, showcasing their exceptional performance in automating mundane tasks \cite{DBLP:journals/corr/abs-2303-18223,DBLP:journals/corr/abs-2304-13712}. LLMs offer an all-in-one solution for versatile user requirements, which marks a step toward the realization of Artificial General Intelligence (AGI).

Despite their impressive abilities, LLMs still struggle to meet users' diverse requirements efficiently and conveniently. First, there is increasing interest in deploying LLMs independently for more private, secure, steady, and fast service \cite{DBLP:journals/corr/abs-2310-07298,DBLP:journals/corr/abs-2312-02003}. Yet, LLMs require tremendous resources for training and deployment, which could not be affordable for average users \cite{DBLP:journals/corr/abs-2302-13971,DBLP:journals/corr/abs-2303-08774,DBLP:journals/corr/abs-2307-09288}. For example, pretraining LLaMA-2-70B takes 1.7 million GPU hours on A100-80G and consumes $2.5\times10^{12}$ Joules of energy, and the inference stage consumes comparable electricity \cite{DBLP:journals/corr/abs-2307-09288, xu2024survey, DBLP:conf/mlsys/WuRGAAMCBHBGGOM22}. Additionally, while user requirements vary individually, LLMs might not always achieve optimal results, especially in specialized areas such as legal, economic, and medical fields. In contrast, tailored small models tend to exhibit superior performance \cite{DBLP:journals/corr/abs-1908-08962, DBLP:journals/corr/abs-2306-11644, DBLP:conf/icml/FuPOSK23, DBLP:conf/acl/HsiehLYNFRKLP23}. However, users might not have adequate expertise or enough time and resources to select and finetune these models, which discourages common users from AI techniques. Therefore, our research aims to \textbf{(I) accommodate the diverse and specific needs of users} and \textbf{(II) simplify the utilization of AI models for the average user.} 

\begin{figure}[!t]
    \centering
    \includegraphics[width=\linewidth]{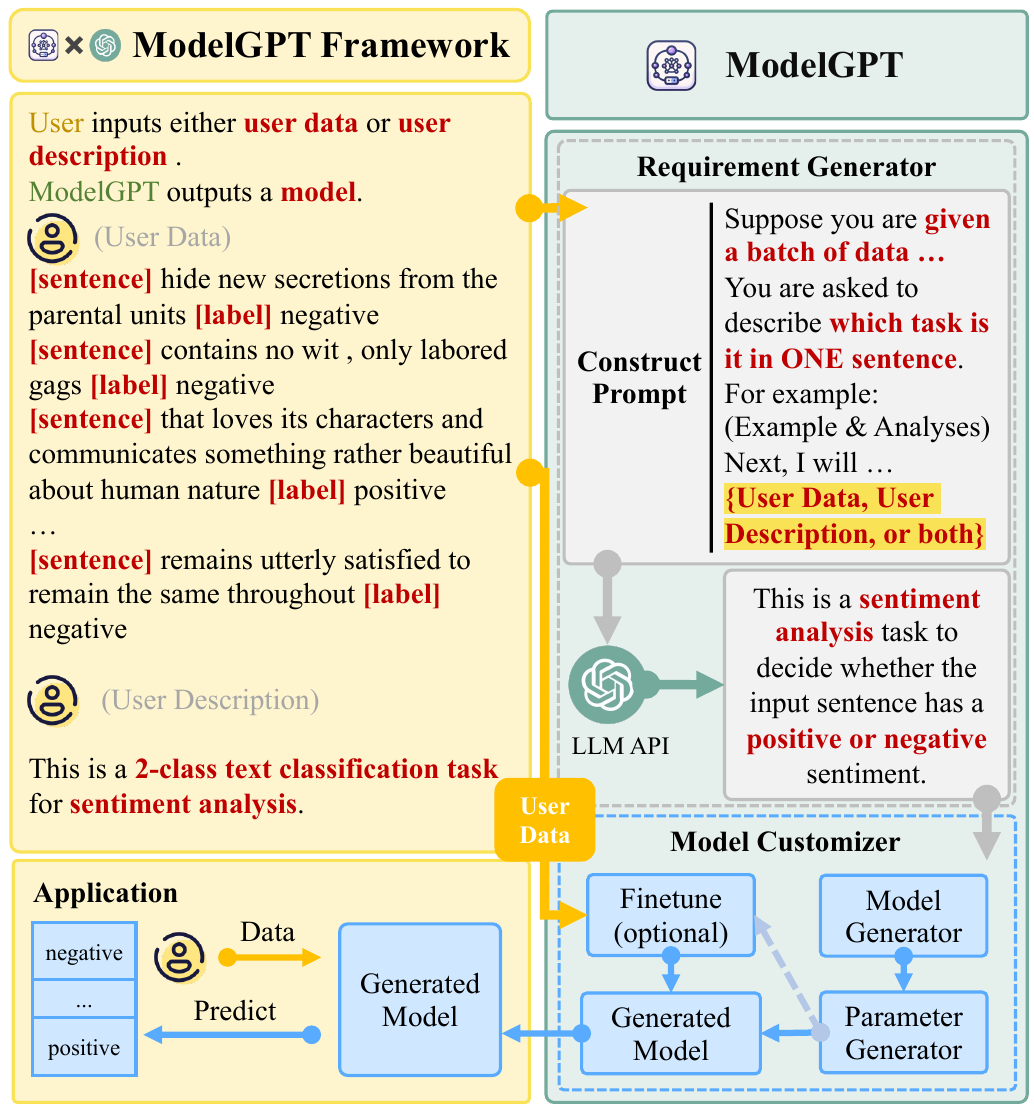}
    \vspace{-0.5cm}
    \caption{Overview of the framework of ModelGPT.}
    \label{fig:intro}
    \vspace{-0.5cm}
\end{figure}

However, these pursuits face certain challenges. First, \textbf{the diversity of user needs necessitates multi-granular model customization}. The difference in requirements would lead to multi-granular changes in the model. While minor changes in user requirements may necessitate adjusting the parameters of the target model for better performance \cite{DBLP:conf/iclr/SagawaKHL20, DBLP:conf/www/LvZ0KWW0SYO023}, significant task alterations might require changing the model's output dimensions or even its architecture. Second, \textbf{to reduce the burden on users, the solution demands both general capabilities (\textit{e.g.} task understanding) and precise customization, preferably without extensive training needs.} Real-world situations often involve limited or no labeled data \cite{DBLP:journals/csur/WangYKN20}, sometimes only a simple task description. Without the ability of task understanding, one might fail to capture user needs precisely. Besides, while large models can efficiently understand and handle general tasks without finetuning, they demand substantial resources for deployment. Conversely, smaller models, although more effective in specific tasks and easy to deploy, require considerable time and expertise for finetuning to achieve optimal performance. Hence, how to provide optimal models easily and efficiently demands further consideration.

In this work, we aim to explore an approach to AGI by generating AI models to efficiently and conveniently meet users' diverse requirements. To address the above challenges, we propose that \textbf{one could leverage the complementary strengths of general large models and specific small models}. Specifically, we could unleash the capabilities of large models to capture user requirements. These requirements can then be used to generate customized small models for users. This framework, which involves using large models to create tailored small models, opens up new possibilities for realizing AGI.

We further instantiate the aforementioned concept, leading to the development of ModelGPT, as depicted in \cref{fig:intro}. Specifically, given users' input (User Data, User Description, or both), ModelGPT constructs a prompt to utilize an LLM for summarizing the task, analyzing data patterns, and formatting it into a user requirement (just a single sentence). This requirement is then encoded into a latent variable and fed to \mb~to determine the most appropriate target model for the given task. Finally, ModelGPT decodes the latent variable into model parameters to get the tailored model, which could be directly used by users for prediction. For instance, as shown in \cref{fig:intro}, for a text classification task of sentiment classification, users can provide data batches (User Data) or just describe the task (User Description). ModelGPT then constructs a prompt, interacts with LLM, and outputs User Requirement in \ma. Next, we generate the model architecture (Distil-BERT base \cite{DBLP:journals/corr/abs-1910-01108}) and its parameters according to the User Requirement. An optional finetune process can be undergone for better performance. Finally, users can predict their data with this tailored model. In short, our contributions can be summarized as follows:
\begin{itemize}[noitemsep,topsep=0pt]
    \item We explore the approach to Artificial General Intelligence to provide tailored off-the-shelf AI models with little data, time, and expertise.
    \item We propose a novel and user-friendly framework ModelGPT, which leverages the ability of LLMs to summarize user requirements and translate them into tailored small models in just ONE forward pass.
    \item We conduct extensive experiments in NLP, CV, and tabular data. ModelGPT can generate tailored models 270x faster than previous methods, while still maintaining comparable performance.
\end{itemize}

\ifthenelse{\equal{\status}{accepted}}{{\color{red}{\textbf{We would like to emphasize that our investigations are still in their initial stages. The present implementation serves as a demonstration of the potential effectiveness of utilizing LLMs in creating tailored models.}}}
}{}

\ifthenelse{\equal{\status}{accepted}}{{
\color{red}{\textbf{We are open to and greatly value discussions and collaborative explorations in this emerging field, and we warmly invite anyone to join us in furthering this research.}}}
}{}

\section{Related Works}
\subsection{Large Language Models}
In recent times, the field of natural language processing (NLP) has been significantly reshaped by the emergence of large language models (LLMs) like ChatGPT \cite{ChatGPT}, GPT-4 \cite{DBLP:journals/corr/abs-2303-08774}, LLaMA \cite{DBLP:journals/corr/abs-2302-13971}, and others. The concept of large language model arises from language model \cite{DBLP:conf/nips/VaswaniSPUJGKP17, DBLP:conf/naacl/DevlinCLT19}, an algorithm used in natural language processing to predict the likelihood of a sequence of words occurring in a sentence. Characterized by deep architectures, billions of parameters, and tremendous training corpus, LLMs have drastically enhanced the ability of machines to understand, interpret, and generate human language \cite{DBLP:journals/corr/abs-2307-06435, DBLP:conf/nips/BrownMRSKDNSSAA20}.

Upon their introduction, LLMs have quickly gained widespread attention and have been applied across various domains, including machine translation, text completion, conversational agents, and so on \cite{shen2023hugginggpt, romera2023mathematical}. However, despite their impressive capabilities, LLMs come with their own set of challenges. Research indicates that in certain specific areas, smaller models can outperform LLMs \cite{DBLP:journals/corr/abs-1908-08962, DBLP:journals/corr/abs-2306-11644, DBLP:conf/icml/FuPOSK23}. Moreover, due to the immense size and complexity of these models, they are often impractical for users to employ or fine-tune, particularly when faced with limitations in computational resources or technical expertise \cite{DBLP:conf/iclr/HuSWALWWC22}.

\subsection{Hypernetworks}
\begin{figure*}[!ht]
    \vspace{-0.1cm}
    \centering
    \includegraphics[width=\textwidth]{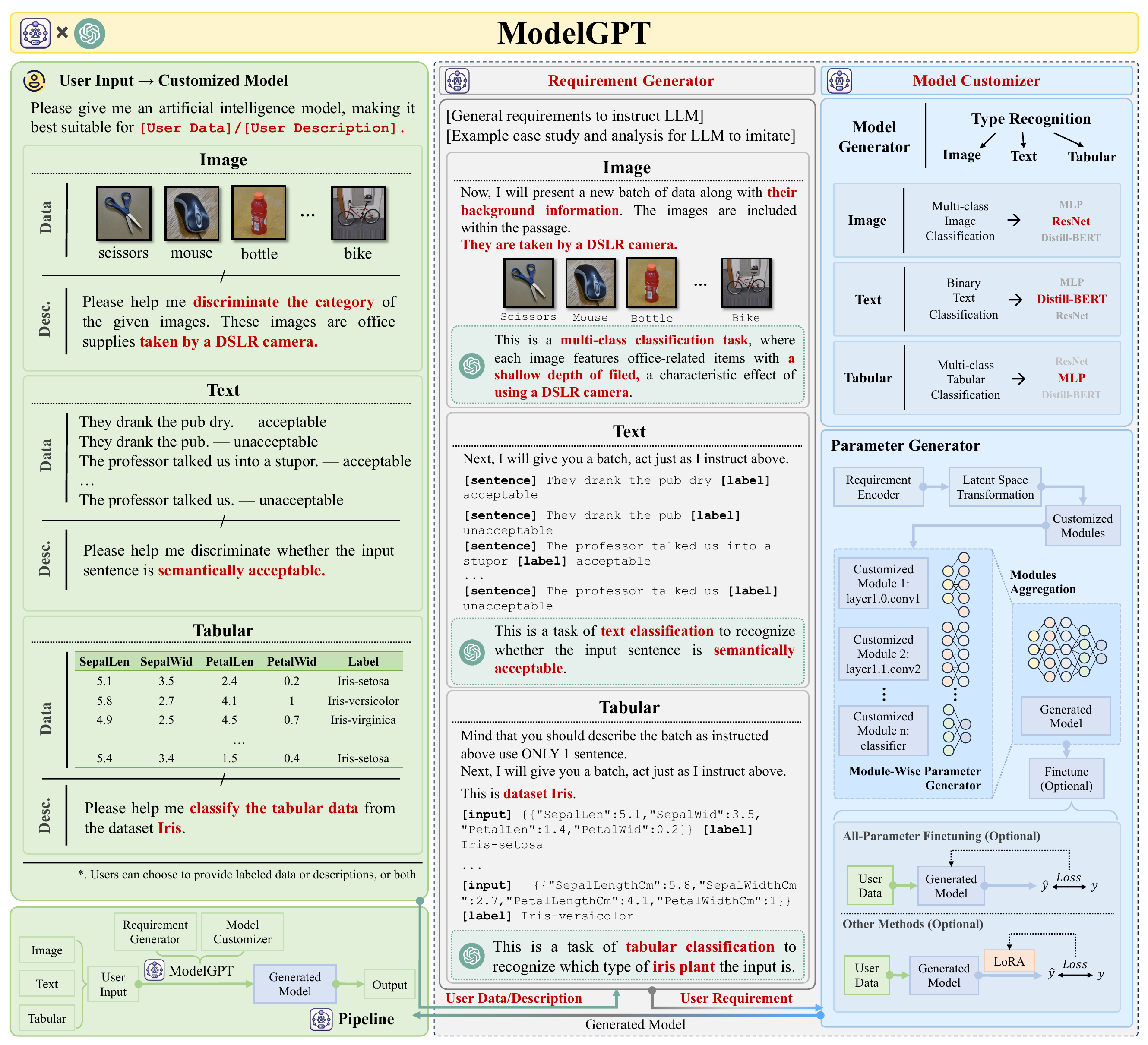}
    \vspace{-0.5cm}
    \caption{Details of the workflow of ModelGPT. Here, we also provide real examples taken from our main experiments.}
    \label{fig:frm}
    \vspace{-0.5cm}
\end{figure*}
Hypernetwork, the model designed to output the weights of another model is first proposed by \citet{DBLP:conf/iclr/HaDL17}. Since it only needs a single forward pass to output model parameters, it provides a fast and efficient alternative to the vanilla pretrain-finetune paradigm. Given its unique capability, it has gained wide attention in various fields like recommendation system, natural language processing, and computer vision. \citet{DBLP:conf/www/LvZ0KWW0SYO023} proposes a framework DUET for efficient device model generalization, which uses hypernetwork to generate the MLP layers of device models for model personalization. \citet{DBLP:conf/acl/IvisonBWHP23} uses hypernetwork to encode task definitions into task-conditioned LoRA adapters \cite{DBLP:conf/iclr/HuSWALWWC22} and applies them to LLMs. \citet{DBLP:conf/cvpr/AlalufTMGB22} proposes HyperStyle, which learns to modulate StyleGAN's weights to faithfully express a given image in editable regions of the latent space. While most hypernetworks are mlps, recently, a few works discuss the potential of hypernetworks with more complex architectures, like GAN \cite{DBLP:conf/icml/RatzlaffL19}, ResNet \cite{DBLP:conf/cvpr/AlalufTMGB22}.

\section{Methodology}
To enable users to generate tailored models with related ease, we propose a comprehensive and novel framework ModelGPT. The workflow of ModelGPT consists of 2 main modules: \ma~and \mb, as depicted in \cref{fig:frm}. \ma~takes User Data or User Description as input and outputs User Requirement, while \mb~translates User Requirement into an off-the-shelf AI model. In this section, we elaborate on the details of our framework.

\subsection{\ma} \label{sec:ma}
Given User Data or User Description, \ma~constructs prompts and utilizes LLM's API to summarize the task, analyze data patterns, and finally format them into one sentence: User Requirement $r$.

Effective prompt design is crucial for accurately distilling patterns from data. On the one hand, User Data might be insufficient, and the lack of data poses challenges to reflect the real distribution in users' scenarios. On the other hand, LLMs tend to highlight simpler patterns that are directly inferable from labels. With the help of User Description, LLM could focus more precisely on the proper and unique patterns. Generally, we summarize the requirements of prompt design as follows:
\begin{itemize}[noitemsep,topsep=0pt]
    \item The type of the task must be pointed out in the final sentence.
    \item Data-specific information, if any, should be reflected in the final sentence, which must only focus on the data itself rather than the labels given.
\end{itemize}

The first one is fundamental: accurately capturing the task's nature, such as image classification or text regression, is essential for identifying an appropriate model.

Conversely, the second one is subtler, focusing on customizing the model to enhance performance. While a general model may perform adequately in standard scenarios, it often struggles with special data patterns like domain shifts in specific user contexts, leading to significant performance drops \cite{Wang_2018, Zhou_2022}. Consequently, \ma~must detect these data patterns present in the data, like spurious correlations between background elements and labels.

We will elaborate on these requirements by case studying in \cref{sec:prompt} and \cref{sec:pd}.

\subsection{\mb}

Given User Requirement, \mb~translates it into a tailored model. It is comprised of 2 sub-modules, \mba~and \mbb, responsible for architecture and parameter generation individually.

\subsubsection{\mba}
Given User Requirement, \mba~determines the architecture of the target model. For example, for a 31-class image classification task, \mba~would use a ResNet-50 model whose classifier's output dimension is 31, while for text regression, \mba~would use a Distil-BERT whose classifier's output dimension is only 1.

\subsubsection{\mbb} \label{sec:mbb}
Once the architecture of the target model $T$ is determined, \mbb~$P(\cdot;\theta_p=(\theta_e,\theta_m,\theta_g))$ generates the parameters $\theta_t$ with User Requirement $r\in R$ as input.

Specifically, $r$ is first encoded by a text encoder $E(\cdot;\theta_e): R\mapsto\mathbb{R}^{l\times d_0}$. $l$ is the pre-set maximum sequence length, while $d_0$ is the size of the hidden dimension of the text encoder. Following previous solutions \cite{DBLP:conf/iclr/WangSMHLB19} to get the sentence embedding $z_0\in \mathbb{R}^{d_0}$, we use the embedding of the first token \texttt{[CLS]} as shown in \cref{eq:z0}:
\begin{equation}
    \vspace{-.1cm}
    z_0 = E(r;\theta_e)[0,:].
    \vspace{-.1cm}
    \label{eq:z0}
\end{equation}
Then, the embedding is fed into a transformation block $M(\cdot;\theta_m):\mathbb{R}^{d_0}\mapsto\mathbb{R}^{d}$, which transforms it into the latent variable $z\in\mathbb{R}^{d}$ as shown in \cref{eq:z}:
\begin{equation}
\vspace{-.1cm}
    z = M(z_0;\theta_m).
    \vspace{-.1cm}
    \label{eq:z}
\end{equation}
Next, we generate the parameters of the target model $\theta_t$ with Module-Wise Parameter Generator $G(\cdot;\theta_g)$ and aggregate them into the model $T(\cdot;\theta_t)$ as shown in \cref{eq:t}:

\begin{equation}
    \theta_t = G(z;\theta_g).
    \label{eq:t}
    \vspace{-.1cm}
\end{equation}

\begin{figure}[!t]
\vspace{-.5cm}
\begin{algorithm}[H]
\caption{Pseudo-code of \mbb~$P(\cdot;\theta_p)$}
\begin{algorithmic}
\REQUIRE $A=\{(D_i=\{X_i,Y_i\},r_i)\}_{i=1}^N$
\ENSURE $\theta_p=(\theta_e,\theta_m,\theta_g)$ satisfies \cref{eq:obj}
\STATE $i\leftarrow1$
\FOR{$\_=0$ to $\sharp$epoch}
\FOR{$(D_i,r_i)$ in $A$}
\FOR{batch in $D_i$}
\STATE Obtain $\theta_t$ with \cref{eq:z0,eq:z,eq:t}
\STATE Use batch to compute the loss and update $\theta_t$
\STATE Compute the difference $\Delta\theta_t$ of $\theta_t$
\STATE Use $\Delta\theta_t$ to compute the gradients of $\theta_p$
\STATE Update $\theta_p$
\ENDFOR
\ENDFOR
\STATE Save best checkpoint according to \cref{eq:obj}
\ENDFOR
\end{algorithmic}
\label{alg:mbb}
\end{algorithm}
\vspace{-.95cm}
\end{figure}

Generally, it is not practical to output the parameters of large-scale models directly for convergence issues \cite{DBLP:conf/cvpr/Dinh0NH22, DBLP:conf/cvpr/AlalufTMGB22}. As a result, we add \textbf{LoRA adapters} \cite{DBLP:conf/iclr/HuSWALWWC22} into the model, generate their parameters, and finally obtain the target model by merging the adapters. For example, as to a certain weight in the target parameter like the weight of the classifier ($a\times b$), $G$ allocates a linear layer that takes the latent variable $z$ as input, outputs the parameters as a $a\cdot b$-dim vector, and then reshapes it into $a\times b$.

The training process of this module is slightly different from the vanilla design, as is demonstrated in \cref{alg:mbb}.

While in application scenes, ModelGPT does not necessarily require users' data, to train ModelGPT, we provide sufficient task-requirement pairs $A=\{(D_i=\{X_i,Y_i\},r_i)\}_{i=1}^N$. The overall process resembles pretraining. Given a certain requirement $r_i$, following the above procedure, we get the parameter ${\theta_t}_i=P(r_i;\theta_p)$ of the target model $T$. With model $T$ and data $D_i=\{X_i,Y_i\}$, we compute the loss of in \cref{eq:loss}. The loss function $l$ here depends on the type of the task. It could be Cross-Entropy for classification tasks and could be MSE for regression.
\begin{equation}
    L_i=\mathbb{E}_{(x,y)\sim D_i}l(T(x;{\theta_t}_i),y).
    \label{eq:loss}
    \vspace{-.1cm}
\end{equation}
While the conventional training process updates the parameter ${\theta_t}_i$ with the computed loss to obtain a better one ${\theta_t}_i'$, we move one step further: we compute the difference $\Delta\theta_{ti}$ of each parameter in ${\theta_t}_i$, use it to compute the gradients of $P(\cdot;\theta_p=(\theta_e,\theta_m,\theta_g))$, and then update our model. To obtain the best performance on all task-requirement pairs, we require the average loss of the pairs to be minimal, as shown in \cref{eq:obj}:
\begin{equation}
\hat{\theta_p}=\arg\min_{\theta_p}\sum_{i=1}^N{L_i}.
    \label{eq:obj}
\end{equation}

Generally, the number of data samples varies from task to task, which could result in an uneven number of task-requirement pairs being generated and in turn lead to inadequate training for tasks with fewer data samples. Such an imbalance is detrimental to the model's overall performance. To address this issue, we manually adjust the portions of each task during the construction of task-requirement pairs.

\begin{table*}[!ht]
\centering
\scriptsize
\vspace{-.3cm}
\caption{Detailed results on GLUE. We use Distil-BERT as the target model. The metric of CoLA and DM is Matthew's Correlation. SST-2, MNLI-m, MNLI-mm, QNLI, RTE, WNLI use accuracy. MRPC and QQP use f1 score and accuracy simultaneously. STS-B uses Pearson-Spearman Correlation. $\sharp$Epoch represents the training epochs for each method to obtain the results. E2E (end-to-end) Runtime measures total task completion time (seconds), while Relative Efficiency scales this runtime against the worst-performing method.}
\setlength\tabcolsep{2.8pt}
\begin{tabularx}{\textwidth}{*{16}{c}}
\toprule
\multicolumn{16}{c}{\bf Results on GLUE Benchmark (Distil-BERT)} \\
\midrule
\bf Methods & \bf CoLA & \bf SST-2 & \bf MRPC & \bf STS-B & \bf QQP & \bf MNLI-m & \mbox{\bf MNLI-mm} & \bf QNLI & \bf RTE & \bf WNLI & \bf DM & \cellcolor{gray!25} \bf Score & \cellcolor{gray!25} \bf $\sharp$Epoch & \cellcolor{gray!25} \bf \makecell{E2E \\ Runtime (s)} & \cellcolor{gray!25} \bf \makecell{Relative \\ Efficiency}\\
\midrule
\bf LoRA &\bf 48.3 & \underline{91.0} & 84.9 / 80.3 & 81.2 / 80.0 & \underline{68.9 / 87.3} & \underline{80.5} & 33.1 & \bf 88.1 & 52.8 & \bf 65.1 & 0.0 & \cellcolor{gray!25} 71.5 & \cellcolor{gray!25} 20 & \cellcolor{gray!25} 75672 & \cellcolor{gray!25} 1.3 \\
\bf Finetune & \underline{45.5} & \bf 91.3 & \bf 86.6 / 80.8 & \bf 82.1 / 80.9 & \bf 69.2 / 87.8 & \bf 81.8 & \bf 80.8 & \underline{87.6} & 56.9 & 63.7 & \bf 35.6 & \cellcolor{gray!25} \bf 74.4 & \cellcolor{gray!25} 20 & \cellcolor{gray!25} 95870 & \cellcolor{gray!25} 1.0 \\
\bf ModelGPT & 39.5 & 88.9 & 85.3 / 78.4 & 80.9 / 80.3 & 63.3 / 83.5 & 77.8 & 78.0 & 84.6 & \underline{69.5} & \underline{64.4} & 28.0 & \cellcolor{gray!25} 73.4 & \cellcolor{gray!25} \bf 0 & \cellcolor{gray!25} \bf 350 & \cellcolor{gray!25} \bf 273.8 \\
\bf ModelGPT-F & 36.9 & 90.8 & \underline{85.5 / 79.4} & \underline{81.3 / 80.5} & 67.0 / 86.6 & 77.8 & \underline{78.1} & 85.8 & \bf 70.0 & 62.3 & \underline{29.9} & \cellcolor{gray!25} \underline{73.8} & \cellcolor{gray!25} \underline{1} & \cellcolor{gray!25} \underline{1101 } & \cellcolor{gray!25} \underline{87.0}\\
\bottomrule
\end{tabularx}
\label{tab:nlp}
\vspace{-.5cm}
\end{table*}

\begin{table*}[!ht]
\centering
\scriptsize
\caption{Detailed results of various methods on 10 tabular classification tasks with accuracy as the evaluation metric. $\sharp$Epoch represents the training epochs for each method to obtain the results. E2E (end-to-end) Runtime measures total task completion time (seconds), while Relative Efficiency scales this runtime against the worst-performing method.}
\setlength\tabcolsep{4.3pt}
\begin{tabularx}{\textwidth}{*{15}{c}}
\toprule
\multicolumn{15}{c}{\bf Results on Tabular Data (MLP)} \\
\midrule
\multirow{2}{*}{\bf Methods} & \multirow{2}{*}{\bf Iris} & \bf Heart & \multirow{2}{*}{\bf Wine} & \multirow{2}{*}{\bf Adult} & \bf Breast & \bf Car & \bf Wine & \bf Dry & \multirow{2}{*}{\bf Rice} & \bf Bank & \cellcolor{gray!25} & \cellcolor{gray!25} & \cellcolor{gray!25} & \cellcolor{gray!25}\\
& & \bf Disease & & & \bf Cancer & \bf Evaluation & \bf Quality & \bf Bean & & \bf Marketing & \multirow{-2}{*}{\cellcolor{gray!25}\bf Average} & \multirow{-2}{*}{\cellcolor{gray!25}\bf $\sharp$Epoch} & \multirow{-2}{*}{\cellcolor{gray!25}\bf \makecell{E2E \\Runtime (s)}} & \multirow{-2}{*}{\cellcolor{gray!25}\bf \makecell{Relative \\ Efficiency}} \\
\midrule
\bf LoRA & \underline{93.3} & \bf 63.0 & 67.3 & 54.7 & \underline{95.9} & \underline{71.3} & 55.0 & \underline{88.9} & 92.5 & 89.8 & \cellcolor{gray!25} 77.2 & \cellcolor{gray!25} 20 &\cellcolor{gray!25} 272 & \cellcolor{gray!25} 1.0 \\
\bf Finetune & 88.9 & 54.3 & \underline{89.1} & \bf 55.2 & \bf 96.5 & 71.0 & \underline{55.3} & \bf 90.6 & \bf 93.1 & \underline{89.9} & \cellcolor{gray!25} 78.4 & \cellcolor{gray!25} 20 & \cellcolor{gray!25} 233 & \cellcolor{gray!25} 1.2 \\
\bf ModelGPT & \bf 100.0 & 60.9 & \bf 94.5 & 54.7 & 95.3 & \bf 71.5 & 54.1 & 85.0 & 92.5 & 89.8 & \cellcolor{gray!25} \underline{79.8} & \cellcolor{gray!25} \bf 0 & \cellcolor{gray!25} \bf 6 & \cellcolor{gray!25} \bf 46.2 \\
\bf ModelGPT-F & \bf 100.0 & \underline{62.0} & \bf 94.5 & \underline{55.1} & \underline{95.9} & \underline{71.3} & \bf 55.4 & 88.8 & \underline{92.9} & \bf 90.0 & \cellcolor{gray!25} \bf 80.6 & \cellcolor{gray!25} \underline{1} & \cellcolor{gray!25} \underline{14} & \cellcolor{gray!25} \underline{20.2} \\
\bottomrule
\end{tabularx}
\label{tab:tabular}
\vspace{-.5cm}
\end{table*}

\begin{table*}[!ht]
\centering
\scriptsize
\caption{Detailed results of various methods on Office-31 (31-class classification). The metric is accuracy, top-3 accuracy, and top-5 accuracy. Our results on Webcam are conducted with no training data provided. $\sharp$Epoch represents the training epochs for each method to obtain the results. E2E (end-to-end) Runtime measures total task completion time (seconds), while Relative Efficiency scales this runtime against the worst-performing method. The average only considers Amazon and DSLR here.}
\setlength\tabcolsep{3.9pt}
\begin{tabularx}{\textwidth}{*{10}{c}|*{3}{c}|*{3}{c}}
\toprule
\multicolumn{16}{c}{\bf Results on Office-31 (ResNet-50, ModelGPT is ZERO-SHOT in Webcam)} \\
\midrule
\bf Domain & \multicolumn{3}{c}{\bf Amazon} & \multicolumn{3}{c}{\bf DSLR} & \multicolumn{3}{c|}{\cellcolor{gray!25} \bf Average} & \multicolumn{3}{c|}{\bf Webcam} & \cellcolor{gray!25} & \cellcolor{gray!25} & \cellcolor{gray!25} \\
\cmidrule{1-13}
\bf Methods & \bf Acc & \bf Acc@3 & \bf Acc@5 & \bf Acc & \bf Acc@3 & \bf Acc@5 & \cellcolor{gray!25} \bf Acc &  \cellcolor{gray!25} \bf Acc@3 & \cellcolor{gray!25} \bf Acc@5 & \bf Acc & \bf Acc@3 & \bf Acc@5 & \cellcolor{gray!25} \bf \multirow{-2}[2]{*}{\makecell{\cellcolor{gray!25} $\sharp$Epoch}} & \cellcolor{gray!25} \bf \multirow{-2}[2]{*}{\makecell{\cellcolor{gray!25} E2E \\ \cellcolor{gray!25} Runtime (s)}} & \cellcolor{gray!25} \bf \multirow{-2}[2]{*}{\makecell{\cellcolor{gray!25} Relative \\ \cellcolor{gray!25} Efficiency}}\\
\midrule
\bf LoRA & 66.4 & 77.7 & \underline{84.8} & 78.4 & 92.2 & \underline{96.1} & \cellcolor{gray!25} 72.4 & \cellcolor{gray!25} 85.0 & \cellcolor{gray!25} 90.5 & 72.5 & 87.5 & \underline{93.8} & \cellcolor{gray!25} 400 & \cellcolor{gray!25} 3393 & \cellcolor{gray!25} 1.1 \\
\bf Finetune & \underline{67.5} & 79.2 & 83.7 & \underline{84.3} & \underline{98.0} & \bf 100.0 & \cellcolor{gray!25} 75.9 & \cellcolor{gray!25} 88.6 & \cellcolor{gray!25} \underline{91.9} & \bf 90.0 & \bf 100.0 & \bf 100.0 & \cellcolor{gray!25} 
 400 & \cellcolor{gray!25} 3770 & \cellcolor{gray!25} 1.0 \\
\bf ModelGPT & 66.4 & \underline{79.9} & 83.7 & \bf 92.2 & \bf 100.0 & \bf 100.0 & \cellcolor{gray!25} \underline{79.3} & \cellcolor{gray!25} \underline{90.0} & \cellcolor{gray!25} \underline{91.9} & 76.2 & 87.5 & 91.2 & \cellcolor{gray!25} 
 \bf 0 & \cellcolor{gray!25} \bf 15 & \cellcolor{gray!25} \bf 257.6 \\
\bf ModelGPT-F & \bf 67.8 & \bf 81.3 & \bf 85.9 & 
\bf 92.2 & \bf 100.0 & \bf 100.0 & \cellcolor{gray!25} \bf 80.0 & \cellcolor{gray!25} \bf 90.7 & \cellcolor{gray!25} \bf 92.8 & \underline{77.5} & \underline{90.0} & 91.3 & \cellcolor{gray!25} \underline{1} & \cellcolor{gray!25} \underline{18} & \cellcolor{gray!25} \underline{206.4} \\
\bottomrule
\end{tabularx}
\label{tab:cv}
\vspace{-.5cm}
\end{table*}

\section{Experiments}
\subsection{Experiment Settings}
To test the efficiency of our proposed framework, we conduct comprehensive experiments in three settings: NLP, CV, and tabular data.

\textbf{NLP } Here, we use GLUE Benchmark \cite{DBLP:conf/iclr/WangSMHLB19}, which has nine sentence- or sentence-pair language understanding tasks built on established existing datasets and selected to cover a diverse range of dataset sizes, text genres, and degrees of difficulty\footnote{https://gluebenchmark.com/}. \mba~chooses Distil-BERT as the target model.

\textbf{Tabular Data } Here, we choose 10 famous tabular classification tasks: Iris \cite{Unwin2021TheID}, Heart Disease \cite{Detrano1989InternationalAO}, Wine \cite{Aeberhard1994ComparativeAO}, Adult \cite{misc_adult_2}, Breast Cancer \cite{Street1993NuclearFE}, Car Evaluation \cite{bohanec1988knowledge}, Wine Quality \cite{Cortez2009ModelingWP}, Dry Bean \cite{Koklu2020MulticlassCO}, Rice \cite{Cnar2019ClassificationOR}, Bank Marketing \cite{Moro2014ADA} from UCI Machine Learning Repository\footnote{https://archive.ics.uci.edu/}. \mba~chooses MLP as the target model.

\textbf{CV } Here, we use Office-31\cite{DBLP:conf/eccv/SaenkoKFD10}, a dataset commonly used in the field of domain adaptation. This dataset contains 31 object categories in three domains: Amazon, DSLR, and Webcam with 2817, 498, and 795 images respectively, different in background, viewpoint, color, etc. \mba~chooses ResNet-50 \cite{DBLP:conf/cvpr/HeZRS16} as the target model. In this experiment, \textbf{both the efficiency and zero-shot ability are tested}. To test the zero-shot ability of our framework, we first train our model with Amazon and DSLR. Then, we directly feed User Requirements extracted by LLM on Webcam's training data to ModelGPT and test the output model on Webcam's test dataset, where ModelGPT sees no Webcam's data but its requirements.

\textbf{Baselines } Since we are the first work in this field, we compare our framework with 2 baselines: \textbf{Finetune} and \textbf{LoRA}. In Finetune, we finetune the target model with all parameters by the training data \textbf{in each task individually}. Since \textbf{our framework uses LoRA to reduce the complexity of our hypernetwork}, we treat fine-tuning the target model with LoRA adapters \textbf{in each task} as the baseline LoRA. The modules trained with LoRA are the same as ModelGPT. In the experiment on tabular data, since the target model MLP is simple, ModelGPT directly outputs its weights, rather than using LoRA adapters. It is important to notice that \textbf{while in the different tasks in one experiment, these baselines give different models, we ONLY need ONE ModelGPT to solve all the tasks in one experiment.}

For our study, we introduce a variant of ModelGPT, designated as ModelGPT-F. This adaptation involves an additional step where, upon ModelGPT's generation of the target model, we perform a single epoch of finetuning involving all parameters, using consistent hyperparameter settings. Detailed hyperparameter configurations are available for reference in \cref{sec:hps}.

\subsection{Results and Observations} \label{sec:r&o}
The results can be seen from \cref{tab:cv,tab:tabular,tab:nlp}, underscoring the efficiency, and exceptional performance of our framework. Here, we provide detailed discussions of our results.

\textbf{ModelGPT yields progressively more significant speed gains as the size of the target model increases.} Leveraging the power of hypernetworks, ModelGPT generates custom model weights in a single forward pass, eliminating the need for a resource-intensive and expertise-dependent training process. This approach yields progressively more significant speed gains over the conventional pretrain-finetune paradigm as the size of the target model increases. The acceleration observed ranges from approximately 40x for a simple MLP (1K) to 270x for Distil-BERT base (66M), marking a 7-fold increase in efficiency. It is also interesting to find out that in the experiments of tabular data, LoRA is much slower than Finetune. The deficiency in speed stems from the target model. Since tabular tasks are rather simple, ModelGPT directly uses multi-layer perceptron, a very shallow neural network isomorphic to LoRA adapters. Hence, directly fine-tuning the target model is more efficient than using LoRA to fine-tune it. These results highlight ModelGPT's exceptional efficacy in producing tailored models, particularly for larger target architectures, demonstrating its potential as a transformative tool in model generation.

\textbf{Inter-task knowledge empowers ModelGPT for enhanced model generation.} Due to hypernetworks' limitations, ModelGPT cannot generate large models directly. Instead, our implementation for sizable target models is to generate LoRA adapters and merge them to construct the final models. This approach may initially seem at most comparable to the baseline LoRA. Yet, in practice, ModelGPT surpasses LoRA in all experiments and even outperforms Finetune in CV and tabular data tasks. This performance boost is largely attributable to the inter-task knowledge gleaned by ModelGPT. Although tasks within a single experiment differ, they share common knowledge. For example, in NLP experiments, both MRPC and QQP tasks focus on semantic equivalence between sentences, and in CV experiments, Amazon, DSLR, and Webcam involve similar classification tasks with unique data-specific characteristics. This observation is also confirmed by our zero-shot success in the Webcam task, where it outperforms LoRA without direct data access, relying solely on User Requirements.

\textbf{ModelGPT not only generates well-performed models but also provides efficient initial weights.} In our main experiments, we introduce a ModelGPT variant that undergoes a single-epoch finetuning of all parameters post-generation. This approach leads to more favorable outcomes, achieving an average performance improvement of 0.8 absolutely while only doubling the total time consumption. We provide detailed analyses of this observation in \cref{sec:init}.

\begin{figure*}[!ht]
    \centering
    \includegraphics[width=\textwidth]{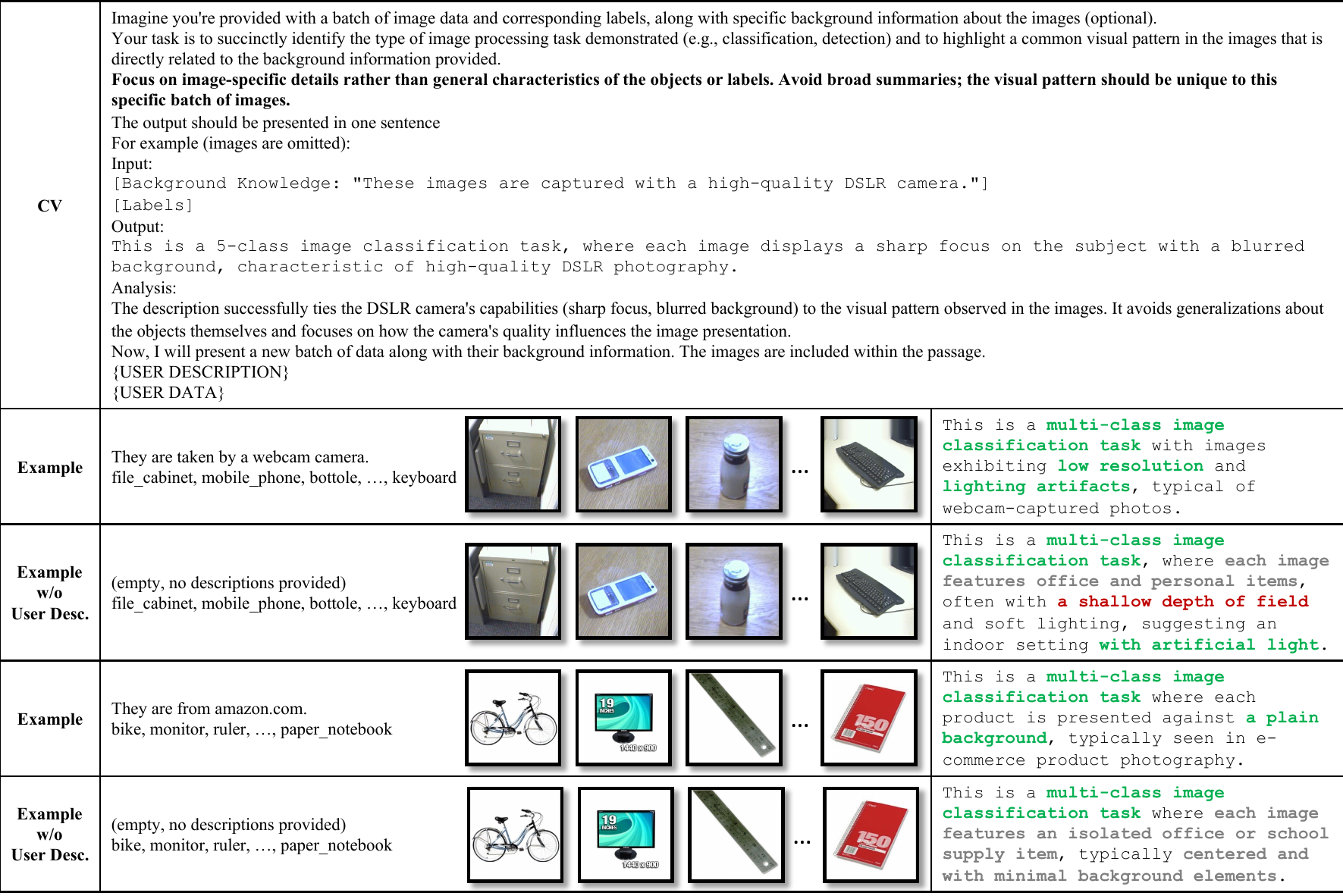}
    \vspace{-0.5cm}
    \caption{Case study on the prompt design. Here, we choose the prompt we use for CV experiments as an example. The first row is the basis of our prompt, leaving the last 2 lines filled with real data and task descriptions. We then provide two pairs of examples in the next rows. In the pair, the only difference between them is whether or not we provide task descriptions. The third column of these examples is the result LLM (GPT-4-vision-preview in this case) outputs. {\color{Green}{\bf The green color texts}} are those reflecting the correct data-specific information, while {\color{Red}{\bf the red ones}} are those reflecting the WRONG information and {\color{Gray}{\bf the gray ones}} are irrelevant information.}
    \label{fig:prompt_s}
    \vspace{-0.5cm}
\end{figure*}

\subsection{More Analyses}
To further testify to the effectiveness of our framework, we provide detailed analyses of ModelGPT's prompt design and the capability of weight initialization.

\subsubsection{Prompt Design} \label{sec:prompt}
Since ModelGPT directly uses LLM's API to summarize users' requirements, it is crucial to design proper prompts for them. As discussed in \cref{sec:ma}, the prompt should briefly tell the type of the task and the data-specific information hidden among the data.

As we stated in \cref{sec:ma}, due to various reasons such as lack of data, summarizing users' requirements often poses challenges. In response, we carefully design the prompt and incorporate users' knowledge (User Data or User Description) into it, resulting in better performance.

We offer a clear case study in our CV experiment, where all the domains are 31-class classification tasks, but each domain has its data-specific information. In Amazon, since the images are from the Amazon website, the images are presented against a white background. In DSLR and Webcam, they are products taken by a DSLR or webcam camera individually. In \cref{sec:pd}, we provide a more detailed analysis of the prompt designing of all the datasets we use in the main experiment.

As shown in \cref{fig:prompt_s}, the first row is the basis of our prompt, which contains our requirement, one example for better reasoning, and the place (last 2 lines) to fill real data.

In the requirement part, we emphasize that LLM ought to succinctly identify the type of image processing task demonstrated (e.g., classification, detection) and point out the data-specific information. To avoid outputting common patterns that can be directly inferred from the labels like these images are office supplies, we reiterate our request in the following sentence, which is bolded in \cref{fig:prompt_s}.

In the example part, we provide an example and analyze it, following the idea of COT (Chain of Thought) \cite{DBLP:conf/nips/Wei0SBIXCLZ22} for better performance.

As we use DSLR domain in the example part of the prompt, we provide results of our prompts on Amazon and Webcam. All the examples successfully point out the type of task. However, without prior knowledge, LLM usually fails to tell the data-specific information.

\begin{figure*}[!th]
\centering
\begin{subfigure}{0.245\textwidth}
    \centering
    \includegraphics[width=.99\linewidth]{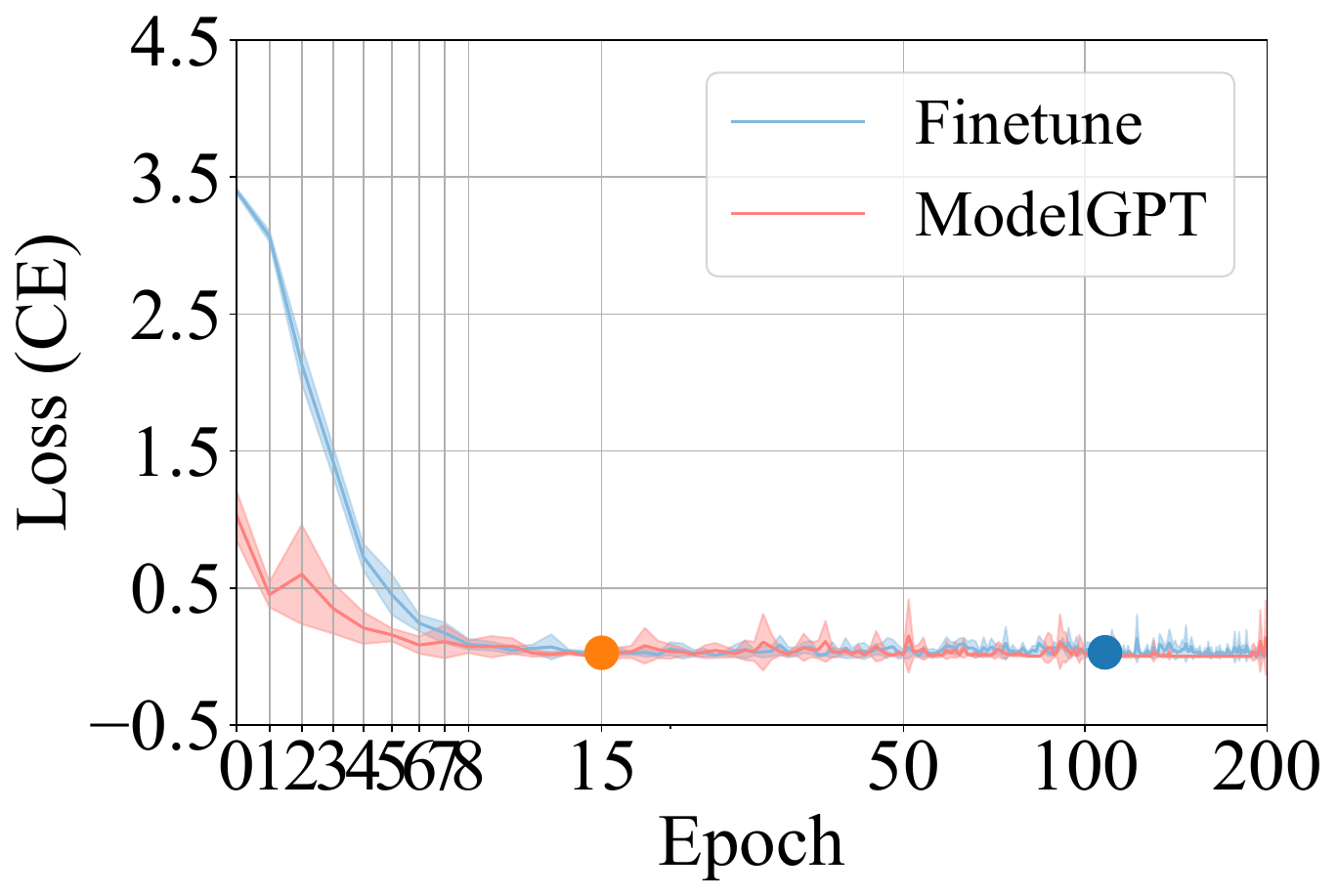}%
    \vspace{-.5\baselineskip}
    \caption{Training Loss}%
    \label{fig:wi0}
\end{subfigure}%
\begin{subfigure}{0.245\textwidth}
    \centering
    \includegraphics[width=.99\linewidth]{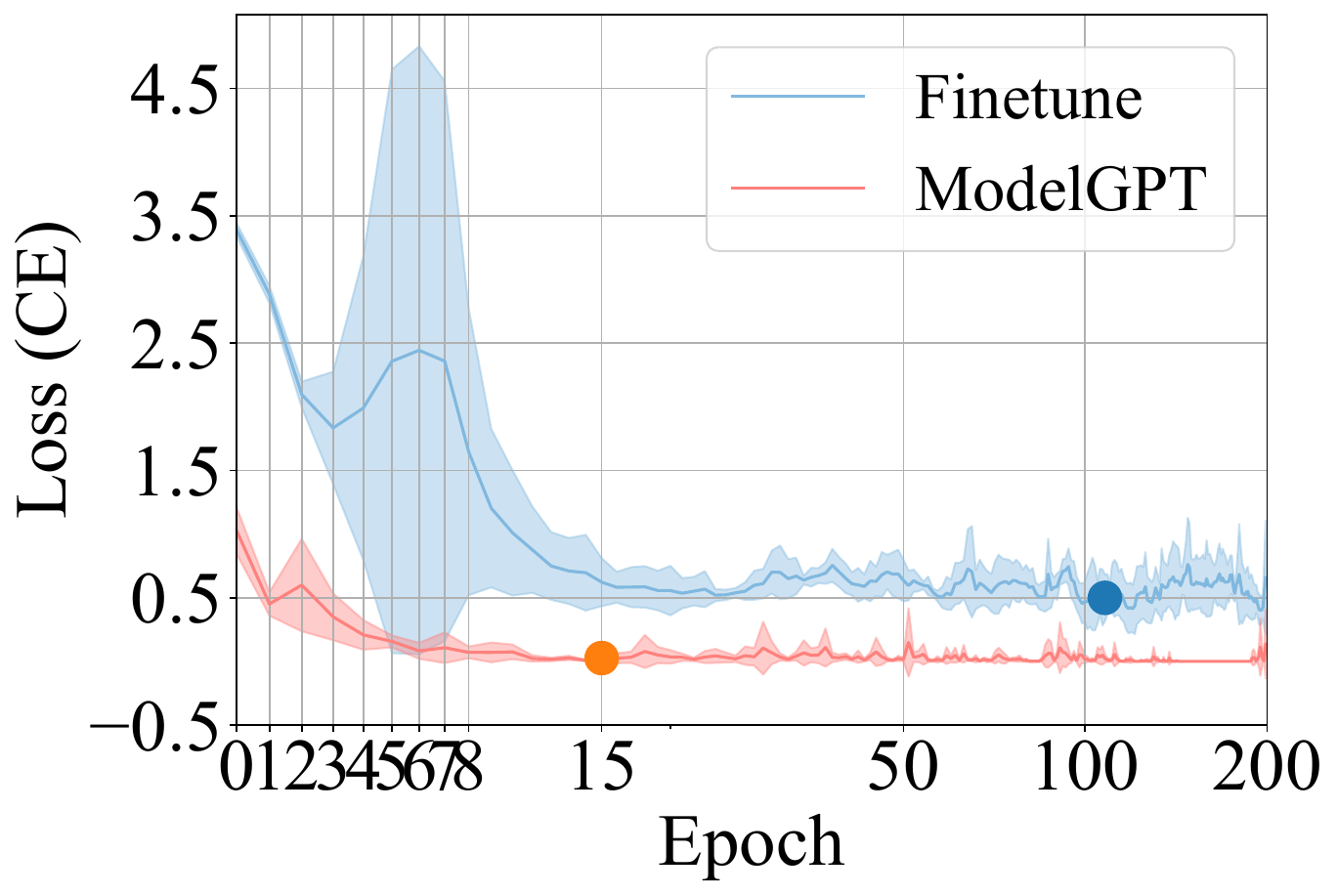}%
    \vspace{-.5\baselineskip}
    \caption{Evaluation Loss}%
    \label{fig:wi2}
\end{subfigure}%
\begin{subfigure}{0.245\textwidth}
    \centering
    \includegraphics[width=.99\linewidth]{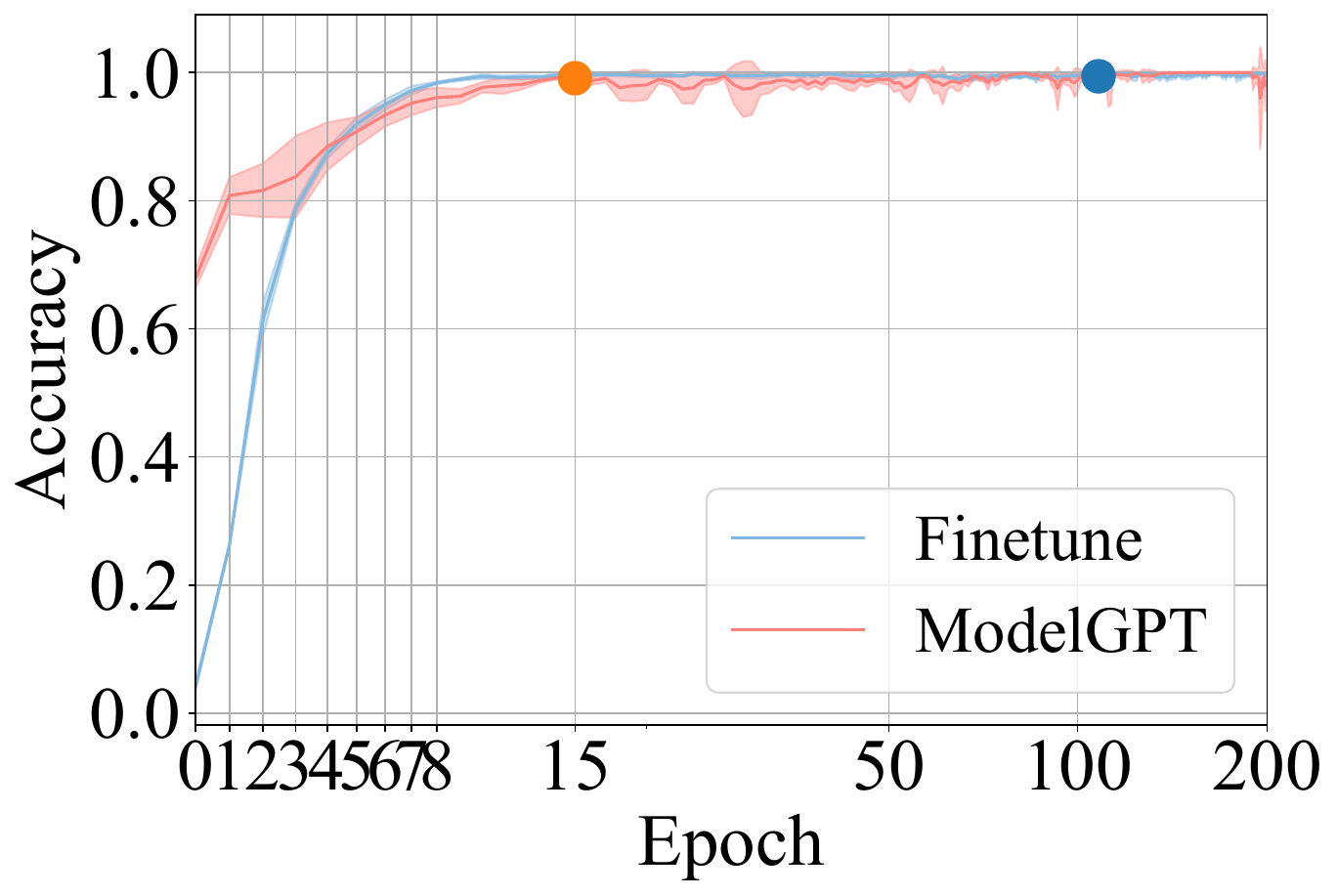}%
    \vspace{-.5\baselineskip}
    \caption{Training Accuracy}%
    \label{fig:wi1}
\end{subfigure}%
\begin{subfigure}{0.245\textwidth}
    \centering
    \includegraphics[width=.99\linewidth]{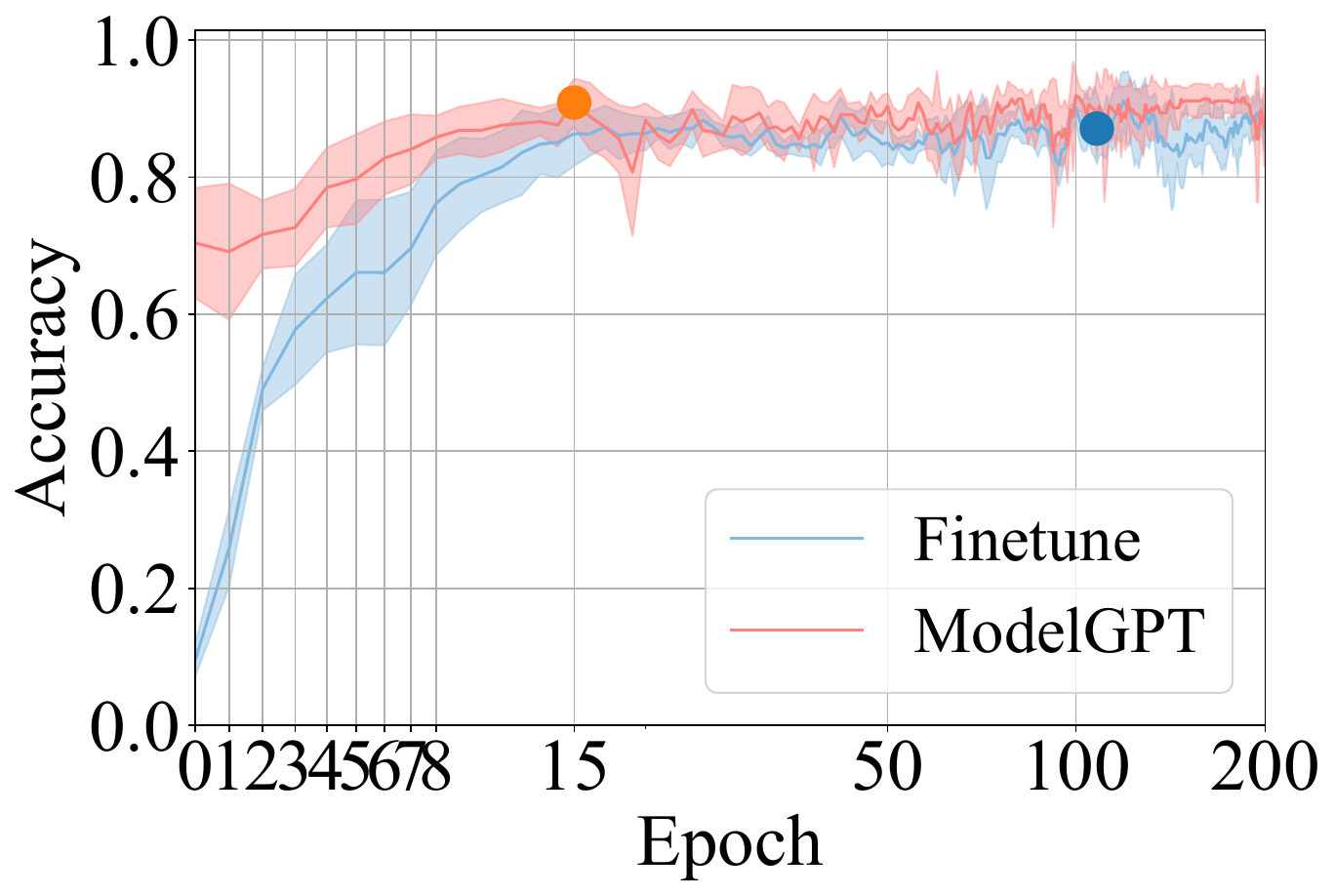}%
    \vspace{-.5\baselineskip}
    \caption{Evaluation Accuracy}%
    \label{fig:wi3}
\end{subfigure}%
\vspace{-1\baselineskip}
\caption{Detailed analyses on the capability of weight initialization of ModelGPT. For clearer comparison, we increase the length of the starting epochs. Meanwhile, we mark the best checkpoint of each method in the figures with a solid round point.}
\label{fig:wi}
\end{figure*}

The first pair of examples are from Webcam, where all the images are taken by a webcam camera. These images are typical of low resolution and light artifacts. User Description we provide is simple: we point out that these images are taken by a webcam camera. With this hint, LLM easily points out the characteristics of this domain. While, in contrast, without this hint, LLM outputs some unimportant or wrong descriptions. To be specific, since data are office items, this information applies to all the domains in Office-31, which is not data-specific information. LLM also mentioned ``a shallow depth of field", which is the pattern of DSLR domain and wrong to Webcam.

The second pair comes from Amazon, where images are directly taken from the website with a plain background. With descriptions provided, LLM precisely catches the data-specific information, while without it, LLM again provides unimportant information.

For more case studies, we refer readers to \cref{sec:pd}.

\subsubsection{Capability of Weight Initialization} \label{sec:init}
In \cref{sec:r&o}, we state that ModelGPT not only generates well-performed models but also provides efficient initial weights. To further testify to this viewpoint, we finetune the target model (ResNet-50) with all parameters under the same hyperparameter setting to Finetune, with the zero-shot output of ModelGPT on Webcam as weight initialization. It is important to notice that the major difference between the two methods lies in the weight initialization. While Finetune uses the weights pretrained on ImageNet \cite{imagenet_cvpr09}, ours uses the weights outputted by ModelGPT, which is generated only under Webcam's task description without any access to Webcam's data.

We save the best checkpoint in evaluation and test it on Webcam's test data. The results, detailed in \cref{tab:init}, reveal a notable aspect: despite a roughly 10\% performance gap compared to Finetune without access to Webcam's data shown in \cref{tab:cv}, our framework exhibits remarkable convergence speed when finetuned with Webcam's training data, using the same hyperparameters as Finetune. Specifically, while Finetune requires 108 epochs to reach optimal evaluation results, our framework, given ModelGPT's zero-shot output as initialization, achieves comparable performance in just 16 epochs, a 6.75-fold increase in speed.

Moreover, we meticulously track the progression of training and evaluation losses, alongside the corresponding accuracy as presented in \cref{fig:wi} with five different seeds. The depicted curves represent the mean value, while the shaded areas denote the range within one standard deviation. All the figures demonstrate the superiority of ModelGPT's output as a weight initialization. As shown in \cref{fig:wi2}, ModelGPT's initialization outperforms the baseline in evaluation throughout the process. Notably, the substantial standard deviation observed in the baseline during the initial epochs in \cref{fig:wi2} can be attributed to the instability often encountered at the onset of training. Moreover, while the baseline shows a marginally improved performance in the later stages of training in \cref{fig:wi1}, our approach demonstrates better performance on evaluation data in \cref{fig:wi3}, suggesting a better generalization capability and robustness.

\begin{table}[!t]
\centering
\scriptsize
\vspace{-.5cm}
\caption{Results on the test dataset using the best evaluation checkpoint of each method. $\sharp$epoch implies the number of epochs for each method to achieve the best evaluation checkpoint.}
\vspace{.05cm}
\begin{tabularx}{\linewidth}{c * {5}{Y}}
\toprule
\multicolumn{5}{c}{\bf Office-31 (Webcam) Results with Different Weight Initializations} \\
\midrule
\bf Methods $\backslash$ Metrics & \bf Acc & \bf Acc@3 & \bf Acc@5 & \cellcolor{gray!25} \bf $\sharp$Epoch \\
\midrule
\bf Finetune & 90.0 & 100.0 & 100.0 & \cellcolor{gray!25} 108 \\
\bf ModelGPT & 95.0 & 98.8 & 100.0 & \cellcolor{gray!25} \bf 16 \\
\bottomrule
\end{tabularx}
\label{tab:init}
\vspace{-.5cm}
\end{table}

\section{Future Work}
While ModelGPT demonstrates good performance in creating customized models, areas for improvement remain. First, the granularity of \mba~can be enhanced. A detailed analysis of factors like task complexity and user resources could lead to more precise model architecture generation. Second, the efficiency of \mbb~needs improvement. Our current method adjusts parameter output dimensions using a dictionary approach, pre-encodes task requirements and parameter shapes, which may not be suitable for broader model generation needs. We plan to address these limitations in our future research.

\section{Conclusion}
In this work, we introduce ModelGPT, a novel and comprehensive framework that harnesses the potential of Large Language Models (LLMs) to meet diverse and specific model generation needs. Emphasizing user-centric design, ModelGPT efficiently transforms data or task descriptions into customized models, catering to the unique requirements of each scenario. This approach not only bridges the gap between complex model generation and user accessibility but also paves the way for a new paradigm in adaptive and efficient model creation. However, we would like to emphasize that our investigations are still in their initial stages. We are open to and greatly value discussions and collaborative explorations in this emerging field.

\section*{Impact Statements}
This paper presents work whose goal is to advance the field of Machine Learning. There are many potential societal consequences of our work, none which we feel must be specifically highlighted here.

\bibliography{ModelGPT}
\bibliographystyle{icml2024}

\newpage
\appendix
\onecolumn
\section{Hyperparameter Setting} \label{sec:hps}
In the main article, we conduct comprehensive experiments on ModelGPT. In this section, we provide detailed hyperparameter settings to reproduce our results in \cref{tab:hps}. Since ModelGPT is a framework that generates target models directly. During the training of ModelGPT (denoted as pretrain for simplicity), we need to train both ModelGPT and the target model to update the overall framework. Therefore, we set their learning rate and weight decay individually. 

\begin{table*}[!h]
\centering
\caption{Detailed Hyperparameter Setting of Our Main Experiments}
\setlength\tabcolsep{15pt}
\begin{tabular*}{\linewidth}{cccc}
\toprule
Parameter $\backslash$ Setting & NLP & CV & Tabular Data \\
\midrule
GPU & A100 80G & A100 80G & A100 80G  \\
\midrule
Optimizer (ModelGPT) & Adam & Adam & Adam \\
Learning Rate (ModelGPT) & 1e-5 & 1e-4 & 1e-3 \\
Weight Decay (ModelGPT) & 1e-4 & 1e-5 & 1e-4 \\
\midrule
Optimizer (Target Model) & Adam & Adam & Adam \\
Learning Rate (Target Model) & 1e-4 & 1e-3 & 2e-2 \\
Weight Decay (Target Model) & 1e-4 & 1e-3 & 1e-4 \\
\midrule
lora\_r & 16 & 8 & NA \\
lora\_alpha & 32 & 16 & NA \\
lora\_droput & 0.05 & 0.1 & NA \\
target\_modules & \texttt{.*[qv]\_lin} & \texttt{layer.$\backslash$..$\backslash$.conv.} & NA \\
\midrule
$\sharp$Epoch (Pretrain) & 50 & 100 & 80 \\
Batch Size & 256 & 256 & 64 \\
Latent Dimension & 768 & 128 & 25 \\
Seed & 2024 & 2024 & 2024 \\
\bottomrule
\end{tabular*}
\label{tab:hps}
\end{table*}

The hyperparameter settings of baselines are similar. We set the number of training epochs to 20, 200, and 20 in NLP, CV, and tabular data individually with the learning rate to be 1e-3, 1e-3, and 2e-2 individually. The LoRA config of baseline LoRA is the same to ModelGPT, except that in tabular data, lora\_r is 4, lora\_alpha is 8, lora\_dropout is 0.1, and target modules is \texttt{mlp$\backslash$.$\backslash$d$\backslash$.*}.

As we stated in \cref{sec:mbb}, to solve convergence issues, we adopt LoRA adapters to the target model and generate their parameters. Besides, since we only need very simple AI models like MLPs to solve tabular tasks, ModelGPT directly outputs target models' parameters rather than LoRA adapters' parameters. The implementation of MLP is shown below. We accordingly mark their hyperparameters concerning LoRA to NA in \cref{tab:hps}.

\begin{verbatim}
class MLP(nn.Module):
    def __init__(self, in_dim: int, out_dim: int,
                 hidden_dim: int, n_layers: int,
                 *args, **kwargs):
        super().__init__(*args, **kwargs)
        self.mlp = nn.Sequential(
            nn.Linear(in_dim, hidden_dim),
            *[
                nn.Linear(hidden_dim, hidden_dim) for _ in range(n_layers)
            ],
            nn.Linear(hidden_dim, out_dim)
        )

    def forward(self, x: Tensor) -> Tensor:
        return self.mlp(x)
\end{verbatim}

\begin{figure*}[!ht]
    \centering
    \includegraphics[width=\textwidth]{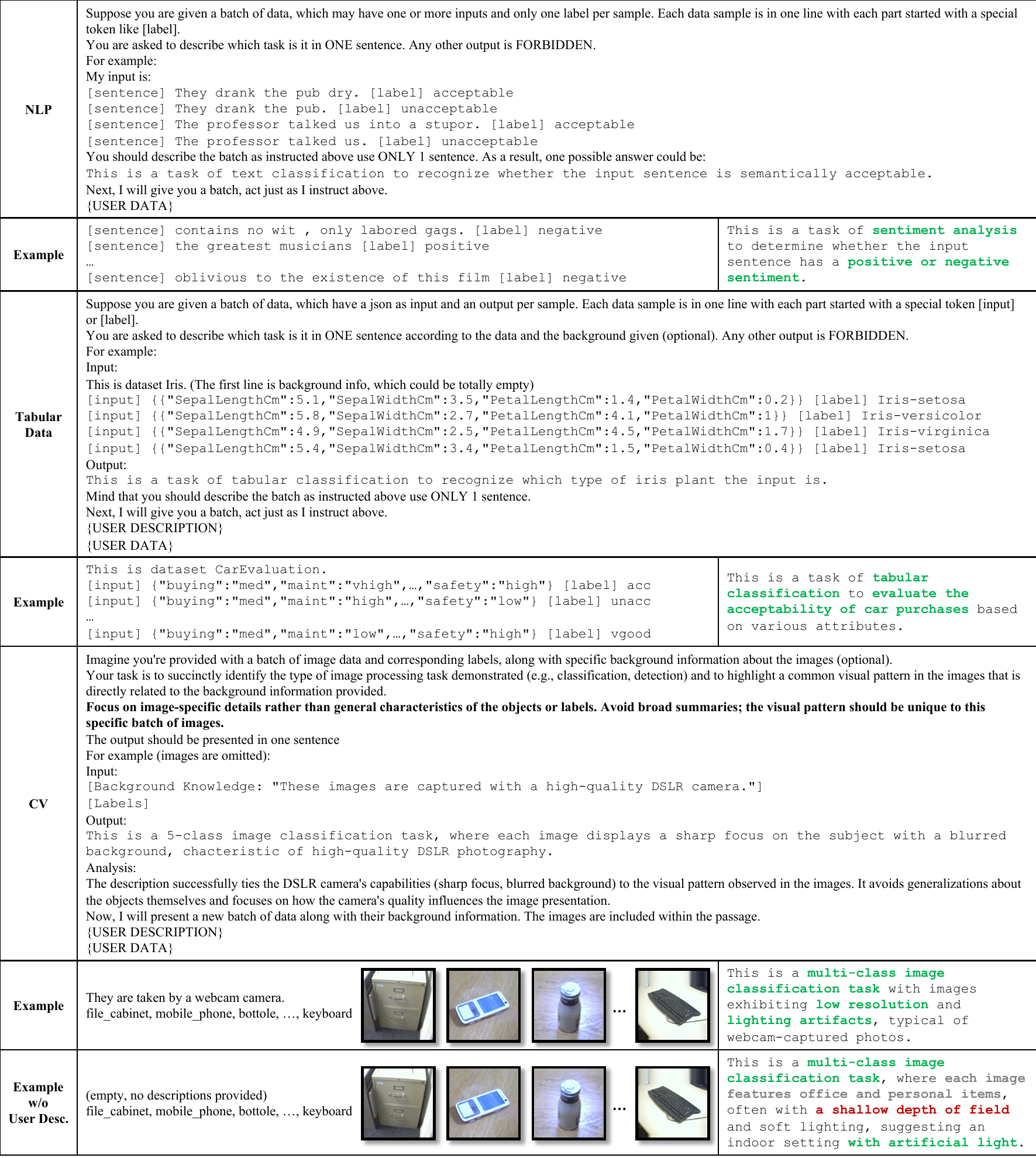}
    \caption{Case study on the prompt design. Here, we provide the prompt template we use in our main experiments on NLP, CV, and tabular data. Each example has 2-3 rows. In each example, the first row is the basis of our prompt, leaving the last 2 lines filled with real data and task descriptions. In CV, we then provide two pairs of examples in the next rows. In the pair, the only difference between them is whether or not we provide task descriptions. The third column of these examples is the result LLM (GPT-4-vision-preview in this case) outputs. {\color{Green}{\bf The green color texts}} are those reflecting the correct data-specific information, while {\color{Red}{\bf the red ones}} are those reflecting the WRONG information and {\color{Gray}{\bf the gray ones}} are irrelevant information.}
    \label{fig:prompt}
\end{figure*}

Generally, in the context of fine-tuning, we could use LoRA adapters to reduce the overall cost. Although such paradigm operates flawlessly in traditional scenarios, it does have some problems in ModelGPT. To be specific, complex models have parameters that are not trained but changed during the stage of finetuning (\textit{e.g.} running mean and variance of BatchNorm Layers). Due to convergence issues, it is not practical for us to generate these parameters alongside the generation of LoRA adapters. However, leaving these modules unsettled would result in unacceptable performance degradation. Hence, to enable ModelGPT to generate complex models, we disable the functionality of these layers in the target model at the expense of less stability with the code below:

\begin{verbatim}
def train(self, mode=True):
    type(model).train.__call__(self, mode)
    for m in self.modules():
        if isinstance(m, nn.BatchNorm2d):
            m.eval()
            m.weight.requires_grad = False
            m.bias.requires_grad = False
            
model.train = functools.partial(train, model)
\end{verbatim}

\section{More Examples about Prompt Design} \label{sec:pd}

In \cref{sec:prompt}, we detailedly discuss the prompt design in our main CV experiments. Here, we also release the prompt templates in our main experiments with other data types. 

As shown in \cref{fig:prompt}, in NLP, we use benchmark GLUE for experiments. The major difference between GLUE's tasks can be directly summarized from their task name since they vary in task type (\textit{e.g.} binary two-input classification, multi-class one-input classification, one-input regression.) As a result, to instruct LLMs to precisely capture user requirements, we just need to figure the task out. In the example, with the prompt given, LLM successfully points out that the given data is a task of binary sentiment analysis. Similar results can be observed in tabular data. Here, different from NLP, we also provide background information (User Description) on the given data. We directly tell LLM that the data comes from datasets such as Iris and Car Evaluation. With User Description and User Data provided, LLM successfully points out that the input is a tabular classification to evaluate the acceptability of car purchases.



\end{document}